\documentclass[a4paper,10pt,twocolumn]{article}
\usepackage[utf8]{inputenc}
\usepackage{multicol,graphicx}
\usepackage{mathptmx}
\usepackage[T1]{fontenc}
\usepackage{textcomp}
\usepackage{url}
\usepackage{amsmath}
\usepackage{amsfonts}
\usepackage{amssymb}
\usepackage{tablefootnote}
\usepackage[T1]{fontenc}
\usepackage[english]{babel}
\usepackage{siunitx}
\usepackage{float}
\usepackage{placeins}

\usepackage[natbib=true, backend=bibtex,style=ieee,doi=false,isbn=false,url=true,maxnames=6,citestyle=numeric-comp,giveninits=true]{biblatex}

\fontfamily{ptm}\selectfont

\bibliography{literature}

\usepackage[font=small]{caption}
\usepackage{subcaption} %

\usepackage{amsmath}

\newcommand{\mysection}[1]{\vspace{0.4cm} \uppercase{#1} \vspace{0.4cm}}
\newcommand{\mysubsection}[1]{\hspace{10pt}\textit{#1:}}

\makeatletter
\newcommand\footnoteref[1]{\protected@xdef\@thefnmark{\ref{#1}}\@footnotemark}
\makeatother

\usepackage{hyperref}

\begin{document}
	
\setlength{\textfloatsep}{10pt plus 1.0pt minus 2.0pt}	
\setlength{\columnsep}{1cm}

\twocolumn[%
\begin{@twocolumnfalse}
\begin{center}
	{\fontsize{14}{18}\selectfont
        \textbf{\uppercase{Uncertainty Quantification for cross-subject Motor Imagery classification}}\\}
    \begin{large}
        \vspace{0.6cm}
        Prithviraj Manivannan$^*$, Ivo Pascal de Jong$^*$, Matias Valdenegro-Toro, Andreea Ioana Sburlea\\
        \vspace{0.6cm}
        Department of Artificial Intelligence, Bernoulli Institute, University of Groningen, The Netherlands\\
        \vspace{0.5cm}
        E-mail: ivo.de.jong@rug.nl
        \vspace{0.4cm}
    \end{large}
\end{center}	
\end{@twocolumnfalse}%
]%

\def\thefootnote{*}\footnotetext{These authors contributed equally to this work}

ABSTRACT: Uncertainty Quantification aims to determine when the prediction from a Machine Learning model is likely to be wrong. Computer Vision research has explored methods for determining epistemic uncertainty (also known as model uncertainty), which should correspond with generalisation error. These methods theoretically allow to predict misclassifications due to inter-subject variability. We applied a variety of Uncertainty Quantification methods to predict misclassifications for a Motor Imagery Brain Computer Interface. Deep Ensembles performed best, both in terms of classification performance and cross-subject Uncertainty Quantification performance. However, we found that standard CNNs with Softmax output performed better than some of the more advanced methods. %

\mysection{Introduction}\label{sec:introduction}

Machine Learning systems for Brain Computer Interfaces (BCI) are normally optimised to their predictive accuracy. The availability of public datasets and benchmarking systems such as the Mother of all BCI Benchmarks \cite{moabb2018} project allow for faster progress in this direction. However, for successful BCI systems there are more aspects that need to be explored. 

This study explores the options of Uncertainty Quantification (UQ) for Machine Learning models \cite{ABDAR2021243} as applied to non-invasive Motor Imagery BCIs. Uncertainty Quantification aims to estimate how likely a prediction from a Machine Learning model is to be correct. For this two types of uncertainty are commonly considered.

\mysubsection{Two types of Uncertainty}
Aleatoric uncertainty (also referred to as data uncertainty) is the uncertainty inherent in the data. This cannot be reduced by better models, only by better EEG recordings or better paradigms. %
Noisy EEG recordings or extracted features that are poorly correlated to the to-be-predicted classes introduce aleatoric uncertainty. 

Epistemic uncertainty (also referred to as model uncertainty) is the uncertainty in the model. This kind of uncertainty can be reduced by collecting more training samples that are similar to what the model is being evaluated on. In BCI contexts this uncertainty can come  from limited amounts of training data \cite{duan2023uncer}, but also from between-subject variability \cite{mcdropout_motorimagery}. %

While there is some Motor Imagery BCI research dedicated to UQ \cite{duan2023uncer, mcdropout_motorimagery, chetkin2023bayesian, milanes2023robust}, it is worth noting that simple methods of estimating aleatoric uncertainty are often readily available. For example, Neural Networks used for classification generally use Softmax or Sigmoid activation functions for the output, which also gives a crude estimate of aleatoric uncertainty. 

This study, like most research on modelling epistemic uncertainty is mostly done in the domain of Deep Learning. %

\mysubsection{Using Uncertainty for Rejection}
UQ is often considered as a method for improving interpretability of predictions from a Machine Learning model \cite{campbell2022robust}. There, the goal is to have a precise and well calibrated prediction of the class probability. This means that a prediction with 90\% certainty should be correct 90\% of the time. This results in methods aimed at addressing overconfidence of Neural Networks \cite{calib_nn}.

However, for BCIs there is often no time for human interpretation of the classification. Instead, the system should automatically deal with certain and uncertain predictions. Typically this means "rejecting" the uncertain predictions and abstaining from sending a control command to the device. We focus on this rejection case, as it aligns with how BCIs are implemented in practice, and highlight that it comes with different methods and metrics.

\mysubsection{Research Aim}
This paper investigates whether UQ methods that account for epistemic uncertainty are able to identify wrong predictions in cross-subject classification. This expands on previous work \cite{mcdropout_motorimagery, milanes2023robust} by exploring a larger variety of UQ methods and by applying a leave-one-subject-out cross validation paradigm to get a more realistic estimate of model performance without calibration.

We investigate whether available UQ methods for CNNs that account for epistemic uncertainty are actually able to reject the uncertain predictions when applied cross-subject better than the crude methods readily available.  

Previous work has shown success with rejection methods \cite{milanes2023robust}, but a comparison with simple baseline methods such as Softmax is missing. Moreover, by using different measures of uncertainty we can see how much aleatoric and epistemic uncertainty contribute to the total uncertainty. Lastly, we cover a wider range of UQ methods and explain how they have different underlying assumptions.

\mysection{Background}

\captionsetup[subfigure]{justification=centering}
\begin{figure}

    \begin{subfigure}{0.5\linewidth}
       \captionsetup{skip=1pt}
      \includegraphics[width=\linewidth]{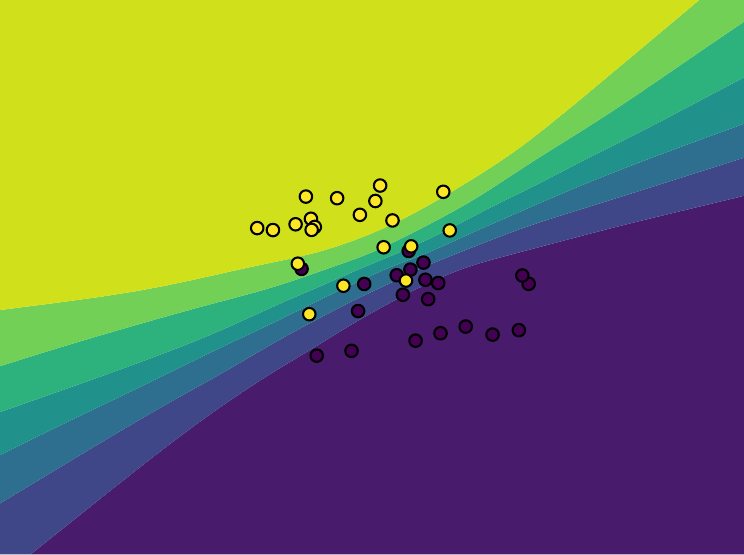}
        \caption{Discriminative model}
        \label{fig:discriminative}
    \end{subfigure}\hfill%
    \begin{subfigure}{.5\linewidth}
           \captionsetup{skip=1pt}

        \includegraphics[width=\linewidth]{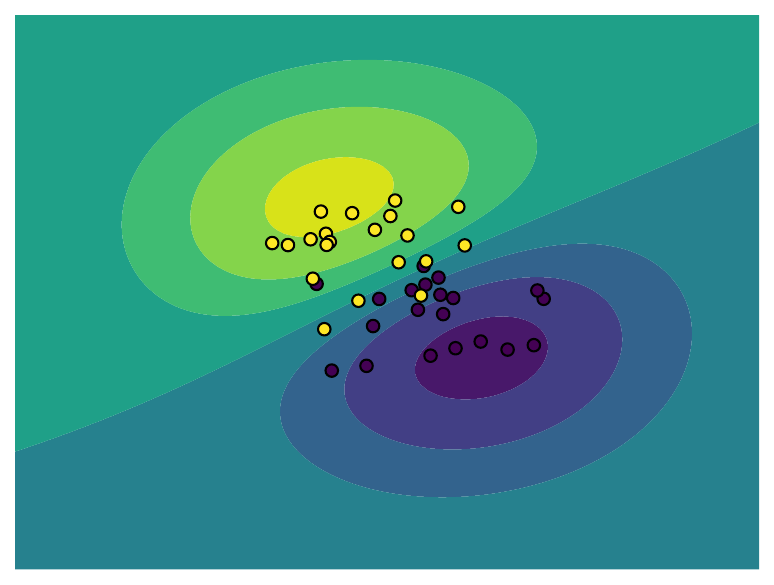}
        \caption{Generative model}
        \label{fig:generative}
    \end{subfigure}\hfill%
    
    \caption{An illustration of a discriminative and a generative model. The yellow and purple dots indicate the training samples of two different classes. The background indicates the prediction. The green color indicates uncertainty.}
\end{figure}

Following \cite{prince2012computer} we consider two assumptions for how epistemic uncertainty may be modelled. \cite{prince2012computer} calls these two assumptions \textit{discriminative} and \textit{generative} models. 

Discriminative models learn a boundary that optimally separates the classes. Samples that are far away from this boundary are considered "certain", whereas samples that are close to this boundary are considered uncertain. When the model uncertainty is considered, these methods consider multiple decision boundaries that are all valid with the training data. When samples fall between different decision boundaries this is considered epistemic uncertainty. Figure \ref{fig:discriminative} shows what this looks like in a 2D feature space. This could be the band power following 2 CSP filters, but a similar concept can also be applied at a higher dimensional space for Neural Networks. 
In contrast, generative models learn the distribution of each class. A sample that matches the distribution of the training data is considered "certain", whereas a sample that is far away from the training data is considered "uncertain". Figure \ref{fig:generative} visualises this concept.

Both approaches have similar behaviour under aleatoric uncertainty. This is seen in the parts where the two classes overlap. However, they exhibit very different behaviour under epistemic uncertainty. Since it is not known which of these underlying assumptions is most suitable it is important to consider models from either family. %

\mysubsection{Bayesian Neural Networks}
Bayesian Neural Networks (BNNs) fall under the category of discriminative models. Standard Neural Networks learn a single optimal vector $\theta$ of the parameters learned on the training data $D$. They then do classification according to the Softmax function to capture aleatoric uncertainty.

BNNs instead consider a weight distribution $p(\theta | D)$. This captures all possible weights for the Neural Networks, based on how well they fit the data. Inference is then made according to the predictive posterior distribution: 
\begin{equation}
    p(y=c|x) = \int \underbrace{p(y=c|x, \theta)}_{aleatoric} \underbrace{p(\theta |D)}_{epistemic} d\theta.
\end{equation}
Truly Bayesian Neural Networks are computationally infeasible, so instead various methods to approximate it have been proposed \cite{de2023uncertainty, ABDAR2021243}. We will be considering MC-Dropout \cite{gal2016dropout}, MC-DropConnect \cite{mobiny2019dropconnect}, Deep Ensembles \cite{lakshminarayanan2017simple} and Flipout \cite{wen2018flipout}.

While they have differences in approximation quality, implementation complexity, and computational cost, they all rely on BNN fundamentals.

\mysubsection{Deterministic Uncertainty Quantification (DUQ)}
DUQ \cite{van2020uncertainty} uses a different approach to Uncertainty Quantification in Neural Networks. DUQ uses a standard Neural Network as a feature extractor, and then learns a centroid for each class. Samples that are far away from the centroids are deemed uncertain, whereas samples that are close to a centroid are deemed certain. 

This different underlying assumption of how uncertainty should arise is inspired by generative models, though DUQ is not actually a generative model. A true generative model models the distribution of the training samples directly, whereas DUQ only models class centroids. Still, this makes it fundamentally different from the discriminative BNNs, and may therefore give different results than the BNN approach. It also means that aleatoric and epistemic uncertainty cannot be clearly distinguished, but they are both included in the predicted uncertainty.

\mysection{Methodology}\label{sec:methodology}

\mysubsection{Dataset}
We used the public Motor Imagery dataset: BCI Competition IV, dataset 2a \cite{bci_dataset}. This dataset contains 22 channel EEG and 3 monopolar EOG channel recordings of 9 subjects performing one of 4 different motor imagery tasks--- left hand (class 1), right hand (class 2), both feet (class 3) and tongue (class 4). %

The sampling rate was 250Hz and the dataset comes pre-applied with a 50Hz notch filter and a bandpass filter of 0.5Hz to 100Hz. 

The Braindecode \cite{braindecode} and MNE \cite{MNE} Python libraries were used to load and pre-process the data.

\begin{figure}
    \centering
    \includegraphics[width=1\linewidth]{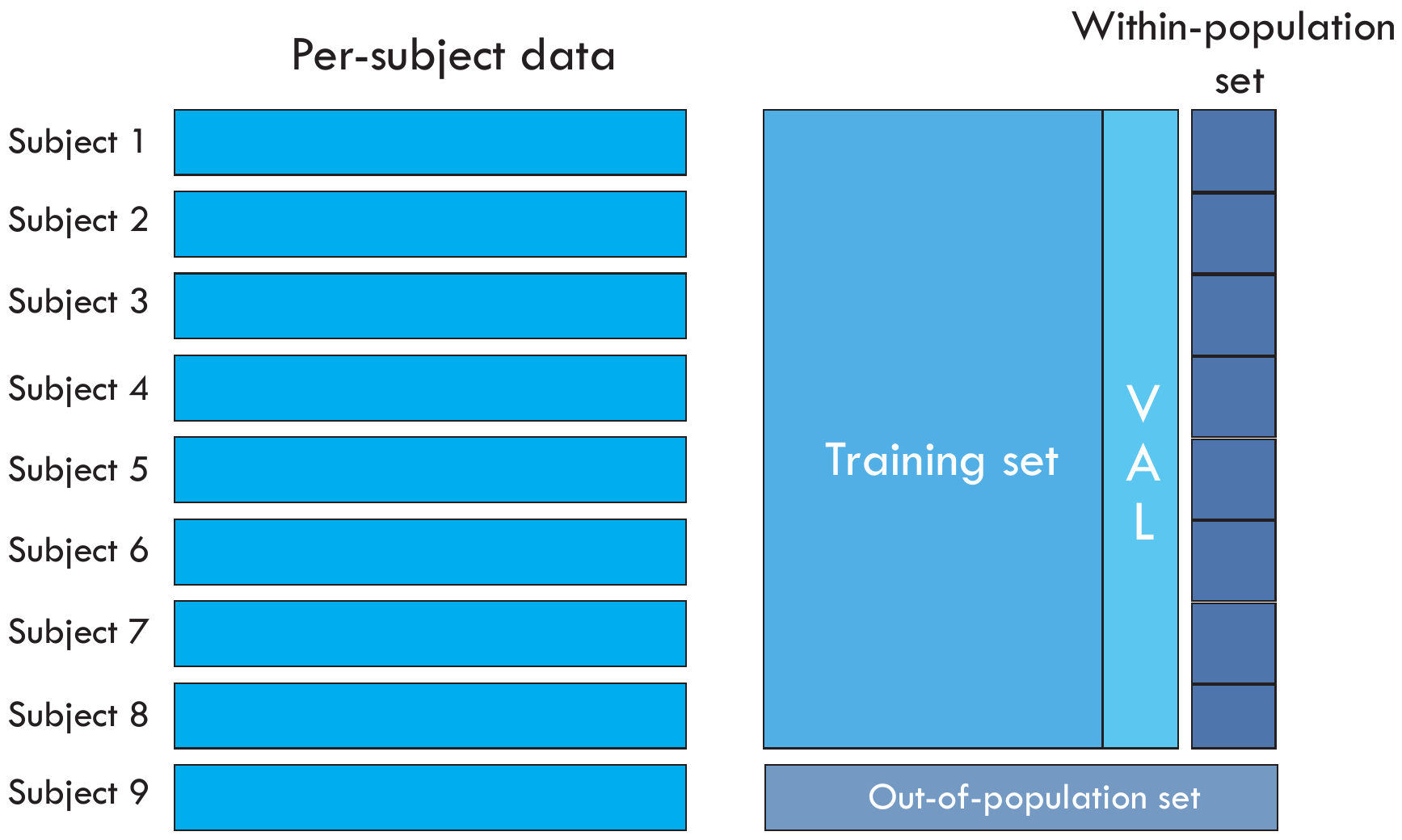}
    \caption{Training setup for a single subject. One subject is excluded and used as an out-of-population set while the other 10\% of the data from each subject is separated into a within-population set. The data of the remaining subjects are concatenated and split 90-10 into a training and validation set. This procedure is repeated for every subject.}
    \label{fig:training_setup}
\end{figure}

The training setup (shown for a single subject in figure \ref{fig:training_setup}) was designed to allow the observation of aleatoric uncertainty and the combination of aleatoric and epistemic uncertainty. This allows the impact of epistemic uncertainty to be observed in isolation. 

We used leave-one-subject-out cross-validation with a slight variation. Normally leave-one-subject-out involves splitting $N-1$ subjects into a training set and leaving the last subject as the out-of-population (cross-subject) set. Our variation to this procedure is as follows:
10\% of the data from each training set subject is used as a within-population test set\footnote{The remaining training data was in turn split into 90\% train and 10\% validation for hyperparameter optimisation}. This within-population dataset allows for an observation with minimal epistemic uncertainty, and comparing it to the cross-subject set allows us to isolate the impact of cross-subject generalisation.

\mysubsection{Preprocessing} \label{ssec:preprocess}
Some EEG pipelines employ extensive signal processing and feature extraction in order to operate with ML algorithms. However, it is often unclear what value each processing step introduces, and various researchers and labs use different pipelines.
The use of CNNs (and DL methods in general) in EEG is promising because of their ability to automatically extract features from raw data and perform classifications, with minimal preprocessing required \cite{tibrewal, no_preprocess}.

Hence the following preprocessing steps are very minimal. It consists of: dropping the EOG channels, converting the EEG signals from volts to microvolts ($\mu V$), applying an exponential moving standardisation with parameters described by \cite{braindecode} and epoching from 0.5 seconds before the trial cue at $t=2s$ to end of the trial at $t=6s$ (for a total trial window of 4.5 seconds). Creating epochs as such leads to a single trial being a matrix $(C, T)$ with $C=22$ being the number of channels and $T=1125$ being the number of timestamps.

\mysubsection{Model Architecture} \label{sec:model_arch}
We used Keras \cite{chollet2015keras} to implement the Shallow ConvNet CNN \cite{braindecode}, and the Keras Uncertainty library \cite{kerasuncertainty} to implement the UQ adaptations. \footnote{All code is available at \url{https://github.com/p-manivannan/UQ-Motor-Imagery}}

Although all UQ methods followed the same Shallow ConvNet architecture, minor differences existed in the implementation of the UQ layers. Two standard models regularised with Dropout and DropConnect were used as baselines.

MC-Dropout and MC-DropConnect and their standard counterparts both had only a single UQ layer. In MC-Dropout this layer was positioned before the dense classification layer with a drop rate of 0.2. In MC-DropConnect it was positioned after the second convolutional layer with a drop rate of 0.1. A grid search was done on a single subject (due to computational complexity) to decide this configuration. Normal Dropout and DropConnect sets the value of a node or weight to 0 during training. The equivalent UQ versions retain this during testing, resulting in slightly different predictions each forward pass, thereby representing epistemic uncertainty.

The Ensemble model simply consisted of 10 standard Shallow ConvNet CNNs, regularised with dropout identical to the dropout baseline model. Disagreement between these 10 models represents epistemic uncertainty. 

Flipout changes the final dense classification layer to a standard dense layer using ReLU activation with 10 units, following which are two flipout layers. Both flipout layers use a prior $P(\theta) = \mathcal{N}(0, 1.0^2) + \pi \mathcal{N}(0, 2.5^2)$ with $\pi = 0.1$. Additionally, the first flipout layer had 10 units. Both sets of parameters were determined using a grid search.

MC-Dropout, MC-Dropconnect and Flipout are stochastic during inference. Therefore, a number of forward passes $T$ needs to be selected. $T$ was chosen to be 50 as it has been found to be point where the improvement in accuracy stabilises  \cite{mcdropout_motorimagery}.

DUQ changes the final layer of the Shallow ConvNet CNN to a dense layer with 100 units using a ReLU activation, following which is an RBF classification layer with a length scale of  $0.4$ with trainable centroids of dimension 100. These parameters were found using a grid search. Additionally, compared to the categorical cross-entropy loss used by the other methods, DUQ utilizes binary cross-entropy.

Other hyperparameters follow common practice in Deep Learning literature. Specifically we set the learning rate ($\num{1e-4}$), loss function (categorical cross entropy), and optimiser (Adam).

\mysubsection{Uncertainty Measures}
The BNN-based methods rely on $T$ forward passes from a stochastic model. Each forward pass predicts class probabilities $p_{c}$, resulting in a distribution over probabilities. To this we can apply various Uncertainty Measures to measure either aleatoric uncertainty, epistemic uncertainty or the total uncertainty \cite{de2023uncertainty}. 

The total uncertainty is based on the mean of the predicted probability for each class and is measured by the Predictive Entropy:

\begin{equation}
    \mathbb{H}_{\text{pred}}(p) = - \sum_c \bar{p}_c \log \bar{p}_c.
\end{equation}

The Expected Entropy first determines the uncertainty of each forward pass, and then takes the average over those uncertainties. 

\begin{equation}
    \mathbb{H_E}(p) = -T^{-1}\sum_t\sum_c p_{ct} \log{p_{ct}}
\end{equation}

In this approach, Expected Entropy takes the "average uncertainty" of each individual model. As such, it only corresponds to aleatoric uncertainty \cite{Mukhoti2021DeterministicNN}. %

Lastly, subtracting the aleatoric uncertainty from the total uncertainty results in the remaining epistemic uncertainty. This measure is referred to as Mutual Information \cite{smith2018understanding}: %

\begin{equation}
    \mathbb{I}(p) \approx \mathbb{H}_\text{pred}(p) - \mathbb{H}_E(p)
\end{equation}

Predictive Entropy and Expected entropy may be applied to a standard Neural Network, but they will result in the same prediction. This approximation for Mutual Information cannot be applied to standard Neural Networks. 

Because DUQ does not follow the same discriminative assumptions for uncertainty, these measures of uncertainty do not apply. Instead, it gives a single uncertainty measure that responds to both aleatoric and epistemic uncertainty.

\mysection{Results}\label{sec:results}

Classification accuracy for each method is given in table \ref{tab:acc}. It can be seen that performance is higher within-population than out-of-population, with Ensembles outperforming all other methods for both groups. The performance of the Ensemble is in-line with other benchmarks for out-of-population and within-population accuracies \cite{no_preprocess} while the other methods are slightly underperforming.

To find out whether UQ can improve performance, uncertainty estimation was treated as a binary classification task, where the aim was to classify wrong predictions as uncertain. Therefore, the Area Under the ROC curve (AUROC) is considered as a performance metric \cite{huang2019evaluating}. Note that this can never approach 100 as the uncertain samples are "guessed", which will be correct 25\% of the time. These would be labelled as false positives in this framework.

This metric is chosen in place of common metrics like Expected Calibration Error \cite{calib_nn}, because our goal is to detect misclassifications, whereas ECE aims to detect overconfidence or underconfidence.

The uncertainty AUROC scores for each method and each uncertainty measure on the within-population set is given in table \ref{tab:lockbox_auroc}. This table shows that Mutual Information (which corresponds only to epistemic uncertainty) performs the worst. Predictive Entropy and Expected Entropy perform similarly, suggesting that the modelling of epistemic uncertainty is not beneficial to the uncertainty estimation. It also shows that DUQ has the worst uncertainty estimation, and that most discriminative models show similar performance.

Table \ref{tab:test_auroc} shows the performance of uncertainty estimation on the out-of-population dataset. The performance of uncertainty estimation is consistently lower here than on the within-population set. Mutual Information, which represents epistemic uncertainty, still does not offer better uncertainty estimation. This suggests that none of the available models are able to fully account for the epistemic uncertainty introduced by cross-subject classification. We again see that DUQ has noticeably worse UQ performance.

It can be seen that the quality of uncertainty estimation is worse cross-population than within-population. This behaviour is inevitable for measures of aleatoric uncertainty, but measures of epistemic uncertainty should be more robust to this \cite{mukhoti2021deep}.

When predictive entropy is disentangled into aleatoric and epistemic uncertainty, it can be seen that epistemic uncertainty based thresholding is consistently slightly worse than aleatoric uncertainty based thresholding. This suggests either that aleatoric uncertainty is more prevalent than epistemic uncertainty, or that epistemic uncertainty is not captured well by the models. Since the accuracy does go down when moving to cross-population, it is clear that there must be an increase in epistemic uncertainty which the models are not accounting for.

\begin{table}[t]
    \begin{small}
    \centering
    \caption{Mean accuracy per subject for each method. Within-population accuracy is higher overall than cross population accuracy, with ensembles outperforming other methods in both categories. Standard DropConnect performs noticeably worse, but most methods perform similar to Standard Dropout. }\label{tab:acc}
    \begin{tabular}{l|l|l} 
        \hline 
        \textbf{Method}&  \textbf{Within pop. Acc\%}& \textbf{Cross pop. Acc \%}\\ \hline 
        Dropout & 68.98 $\pm$ 2.73 & 55.54 $\pm$ 7.95  \\
        MC-Dropout & 69.00 $\pm$ 2.73 & 55.56 $\pm$ 7.94  \\
        DropConnect & 66.67 $\pm$ 2.23 & 53.51 $\pm$ 11.67 \\ 
        MC-DropConnect &  69.27 $\pm$ 1.34  & 54.96 $\pm$ 9.76\\ 
        Flipout & 69.90 $\pm$ 2.55 & 54.99 $\pm$ 8.67 \\
        Ensembles & \textbf{73.05 $\pm$ 2.22}  & \textbf{59.05 $\pm$ 8.11} \\
        DUQ & 70.47 $\pm$ 2.93 &  55.42 $\pm$ 9.16\\ \hline
    \end{tabular}

    \end{small}
\end{table}

\begin{table*}[]
    \centering
    \begin{small}
    \caption{Uncertainty AUROC scores for each method both within-population and out-of-population. Predictive Entropy and Expected Entropy both perform equally well for all BNN models. At the same time Mutual Information performs noticeably worse, and shows more difference for the different models. The uncertainty of all models and all uncertainty measures is consistently worse when moving out-of-population.}\label{tab:lockbox_auroc}
    \captionsetup[subtable]{justification=centering}

    \begin{subtable}{\textwidth}
    \centering
    \caption{Within-population}\label{tab:lockbox_auroc}

    \begin{tabular}{l|l|l|l}
        \hline 
        \textbf{Method}&  \textbf{Predictive Entropy (Ale+Epi)}& \textbf{Expected Entropy (Ale)} & \textbf{Mutual Information (Epi)}\\ 
        \hline 
        Standard Dropout & 76.07 $\pm$ 2.918& 76.07 $\pm$ 2.918& - \\
        MC-Dropout & 76.07 $\pm$ 2.927& 76.06 $\pm$ 2.927& 74.24 $\pm$ 3.398\\
        Standard DropConnect & 75.44 $\pm$ 2.691& 75.44 $\pm$ 2.691& - \\ 
        MC-DropConnect & 75.3 $\pm$ 3.303& 75.29 $\pm$ 3.297& 73.33 $\pm$ 2.835\\
        Flipout & 75.56 $\pm$ 2.461& 76.70 $\pm$ 2.460& 70.56 $\pm$ 2.488\\
        Ensembles &  76.92 $\pm$ 2.868& 76.66 $\pm$ 3.046& 70.02 $\pm$ 2.064\\
        DUQ & 73.19 $\pm$ 2.379& - & - \\

    \end{tabular}
    \end{subtable}

    \begin{subtable}{\textwidth}
    \centering
    \caption{Out-of-population}\label{tab:test_auroc}

    \begin{tabular}{l|l|l|l} 
        \hline 
        \textbf{Method}&  \textbf{Predictive Entropy (Ale+Epi)}& \textbf{Expected Entropy (Ale)} & \textbf{Mutual Information (Epi)}\\ 
        \hline
        Standard Dropout & 67.46 $\pm$ 4.646& 67.46 $\pm$ 4.646&  - \\
        MC-Dropout & 67.43 $\pm$ 4.611& 67.43 $\pm$ 4.611& 66.6 $\pm$ 4.164\\
        Standard DropConnect & 68.23 $\pm$ 4.532& 68.23 $\pm$ 4.532& - \\ 
        MC-DropConnect & 68.48 $\pm$ 4.625& 68.48 $\pm$ 4.626& 66.82 $\pm$ 5.311\\
        
        Flipout & 67.79 $\pm$ 5.156& 67.79 $\pm$ 5.152& 63.95 $\pm$ 4.024\\
        Ensembles & 67.39 $\pm$ 5.446& 67.29 $\pm$ 5.564& 63.86 $\pm$ 4.354\\
        DUQ & 65.30 $\pm$ 4.01&  - & - \\ \hline
    \end{tabular}

    \end{subtable}
    \end{small}
\end{table*}

It can be seen that no BNN method is substantially better than another at uncertainty quantification. Only DUQ performs substantially worse than other methods, performing even lower than standard neural networks. %

\mysection{Discussion and Conclusions}\label{sec:conclusions}

Surprisingly, we find that the specific UQ methods designed to observe epistemic uncertainty are not able give better uncertainty estimations than a similar Neural Network with Softmax activation. It is still possible for all methods to reject some of the uncertain sample to increase accuracy, but this is trivial. 

A possible reason for this is that since aleatoric uncertainty seems more prevalent, the ability of these UQ methods to take into account epistemic uncertainty does not help, hence explaining how standard models are able to achieve comparable performance. However, it is clear that the decrease in accuracy should be attributable to an increase in epistemic uncertainty. This could be caused by how these methods model uncertainty, but the results show that the discriminative models and DUQ suffer the same problems.

Overall we can conclude that adding UQ methods did not result in better uncertainty estimation than standard (softmax) models for the purposes of rejecting difficult samples cross-subject.

\mysubsection{Relation to background}
Our findings contradict the expectation that cross-subject classification should introduce epistemic uncertainty, and that therefore BNN should perform better. 

Epistemic uncertainty should arise when a model is tested on data that is different from the data it was trained on. In this case, the cross-subject testing samples are different from the data that the model is trained on, but the models capturing epistemic uncertainty were not able to offer better uncertainty estimates. 

It is difficult to attribute this to problems with a specific approximation of BNNs, as a variety of approximations show this effect consistently. We also cannot attribute this to flaws in the discriminative model as shown in Figure \ref{fig:discriminative}, because this problem is consistent even when using DUQ which has a fundamentally different assumption of uncertainty. 

The previous studies in this direction \cite{mcdropout_motorimagery, milanes2023robust} show more positive findings for approximations of BNN, but by considering an equivalent CNN and using Softmax as a baseline we find that BNNs are not universally beneficial.

\mysubsection{Limitations}
Our study also only focuses on the use of uncertainty for rejecting difficult samples, and does not actively look at the absolute epistemic uncertainty. It may be that the epistemic uncertainty did increase for cross-subject samples, but if this happens uniformly for a given subject we are not able to capture it. This does not affect the validity of the findings, but does make it harder to know why these Bayesian Neural Networks are not performing well. 

There may also be limitations underlying how Predictive Uncertainty is disentangled into aleatoric and epistemic uncertainty. The proposed approach follows a line of existing work \cite{smith2018understanding, Mukhoti2021DeterministicNN}, but there is also a line of work that assumes an entirely different formulation for disentangling uncertainty \cite{valdenegro2022deeper, kendall2017uncertainties}. There, the BNNs have two outputs. One for predicting the prediction, and one for the variance. The mean of the variances is then the aleatoric uncertainty, and the variance of the predictions is then epistemic uncertainty. This approach explicitly models aleatoric and epistemic as part of the model, which may give more favourable results.

\mysubsection{Directions for future research} We showed that UQ did not work to reject the cross-subject samples with the most epistemic uncertainty. However, it may still be usable for deciding whether or not to make a prediction under noisy EEG, or for identifying a model well suited for a certain subject, or even for detecting off-task thoughts.

We want to emphasise the need to study the behaviour and uses of uncertainty estimates from non-Deep Learning models. Classical Machine Learning models for classification often come with an adaptation to return class probabilities, but the behaviour of these may vary substantially. Assessing their ability to reject segments of EEG that are likely to be false positives may allow for more robust BCI systems. 
The robustness granted by good UQ may be a step towards making BCIs more usable outside of the lab.

\mysection{references}

\printbibliography[heading=none]

\end{document}